\documentclass[conference]{IEEEtran}
\IEEEoverridecommandlockouts
\usepackage{cite}
\usepackage{amsmath,amssymb,amsfonts}
\usepackage{algorithmic}
\usepackage{graphicx}
\usepackage{textcomp}
\usepackage{colortbl}
\usepackage{url}
\usepackage{tabularray}
\usepackage{multirow}
\usepackage{stfloats}
\usepackage{pifont}
\usepackage{bm}
\usepackage{float}
\usepackage{hhline}
\usepackage{booktabs}
\usepackage{makecell}
\floatstyle{plaintop}
\restylefloat{table}
\usepackage[pagebackref,breaklinks,colorlinks]{hyperref}
\usepackage[capitalize]{cleveref}
\crefname{section}{Sec.}{Secs.}
\Crefname{section}{Section}{Sections}
\Crefname{table}{Table}{Tables}
\crefname{table}{Tab.}{Tabs.}

\usepackage[table,xcdraw]{xcolor}
\def\BibTeX{{\rm B\kern-.05em{\sc i\kern-.025em b}\kern-.08em
    T\kern-.1667em\lower.7ex\hbox{E}\kern-.125emX}}
\begin{document}

\title{Prototype Learning-Based Few-Shot Segmentation for Low-Light Crack on Concrete Structures\\

\thanks{Identify applicable funding agency here. If none, delete this.}
}

\author{
    \IEEEauthorblockN{1\textsuperscript{st} Yulun Guo}
~\\
}

\maketitle

\begin{abstract}
    Cracks are critical indicators for evaluating the structural safety of concrete infrastructure. 
    The application of computer vision techniques for automated and non-destructive surface crack detection of surface cracks holds significant engineering importance for preventing structural failure. 
    However, in real-world scenarios, cracks often occur in poorly illuminated environments such as tunnels and bridge undersides. 
    Low-light conditions obscure the distinction between foreground and background regions in images, thereby degrading the accuracy of computer vision–based crack segmentation. 
    Moreover, pixel-level annotation of cracks in low-light images is extremely time-consuming and labor-intensive. 
    Most existing deep learning methods rely on large-scale and well-illuminated datasets, which makes crack segmentation under low light conditions particularly challenging.
    To address these challenges, we propose an improved dual-branch prototype learning network that integrates Retinex theory with few-shot learning for low-light crack segmentation.
    The reflectance components derived from Retinex theory guides the network to learn illumination-invariant global representations, while the metric learning strategy reduces the dependence on large-scale annotated low-light crack datasets. 
    Furthermore, we design a cross-similarity prior mask generation module that computes high-dimensional similarities between query and support features, producing a prior mask that captures crack location and structural information. 
    In addition, we design a multi-scale feature enhancement module to fuse multi-scale features with the prior mask, thereby alleviating spatial inconsistency in few-shot learning.
    Extensive experiments on multiple benchmark datasets demonstrate that the proposed method consistently outperforms existing state-of-the-art approaches under low-light conditions. 
    The code is available at \url{https://github.com/YulunGuo/CrackFSS}.
\end{abstract}


\section{Introduction}
Cracks are the most common defects in critical infrastructure, such as bridges, road pavements and tunnel linings.
Their occurrence often indicates potential degradation in structural performance and serviceability~\cite{yan2022cycleadc}.
Consequently, regular and automated crack detection is essential for evaluating damage severity and ensuring the long-term safety of infrastructure systems~\cite{nguyen2023deep}.

With the continuous advancement of computer vision technologies, non-destructive crack detection based on image acquisition and digital image processing has become mainstream in structural inspection engineering. 
Compared with traditional digital image processing methods that rely on hand-crafted features, deep learning methods automatically learn discriminative crack representations through a data-driven approach, improving detection accuracy and robustness in different scenarios~\cite{nguyen2023deep},\cite{yang2022datasets}.

\begin{figure}[!t]
    \centering
    \includegraphics[width=0.5\textwidth]{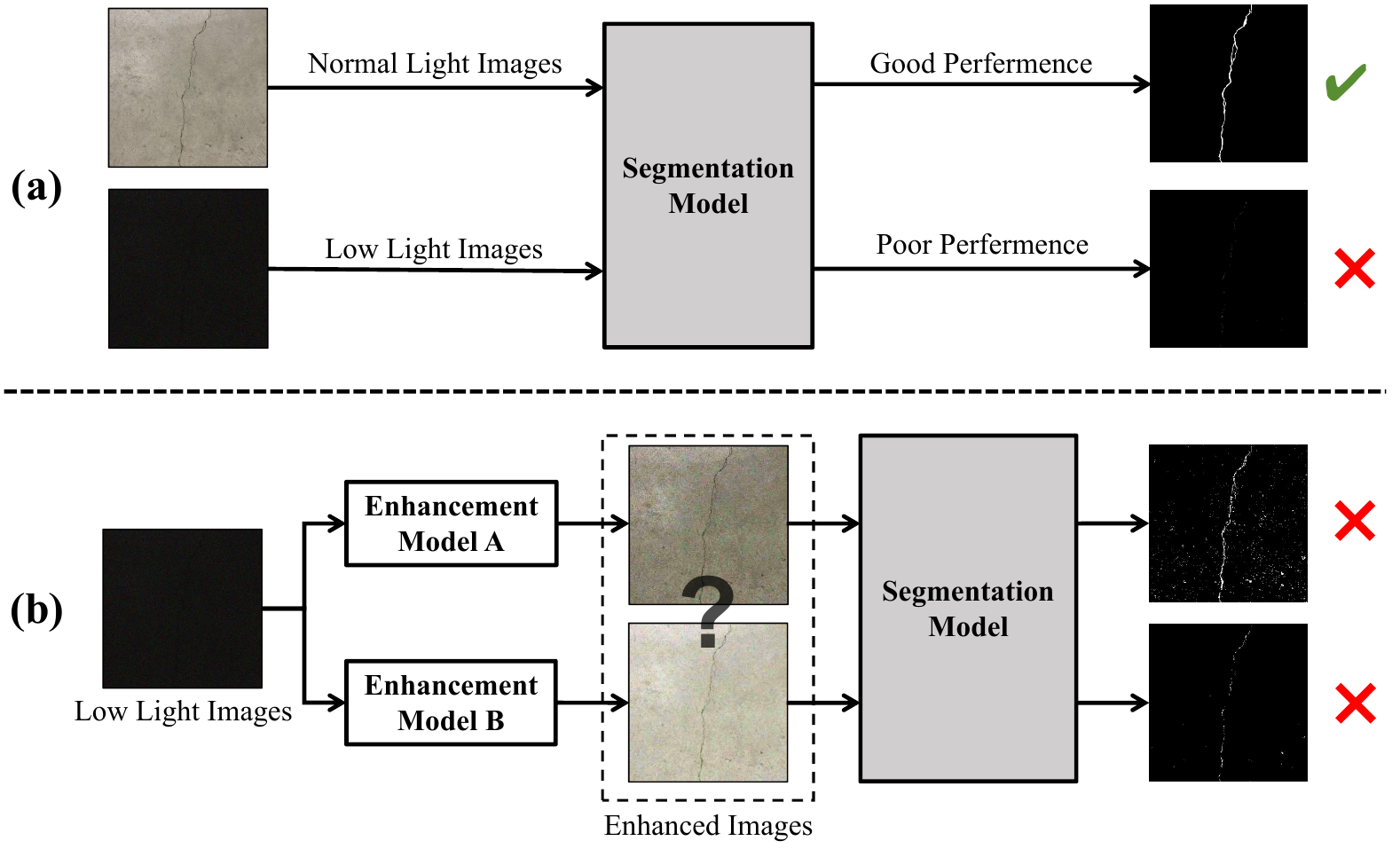}
    \caption{
            Different strategies for low-light crack segmentation:  
            (a) Directly applying the same segmentation model to images captured under varying illumination conditions results in inconsistent segmentation performance; 
            (b) Employing different image enhancement models introduces additional uncertainty in the segmentation results produced by the same segmentation model.
            }
    \label{fig:image1}
\end{figure}

However, as illustrated in Fig.~\ref{fig:image1}, crack segmentation in low-light environments poses significant challenges to existing deep learning methods~\cite{wang2022automatic,fan2023pavement,li2019crack}. 
Images captured under low-light conditions typically suffer from low contrast, severe noise and blurred crack boundaries, which leads to significant performance degradation when segmentation is performed directly. 
Consequently, most low-light crack segmentation pipelines commonly adopt a two-stage framework that integrates image enhancement and segmentation.
Nevertheless, the effectiveness of the low-light enhancement stage has a critical impact on the accuracy of subsequent segmentation. 
Moreover, the acquisition and pixel-level annotation of low-light crack images are costly, and the limited availability of such samples prevents deep learning models from learning robust and discriminative feature representations.

To address the challenges posed by illumination variations and limited supervised training samples, this paper proposes a dual-branch prototype network based on Retinex theory inspired by CrackNex~\cite{CrackNex}. 
The proposed network facilitates the learning of global illumination invariance by employing a dual-branch architecture to process RGB images and corresponding reflectance components in parallel. 
To further enhance the network’s ability to capture crack structural and positional information, we design a Cross-Similarity Prior Mask Generation (CSPMG) module to establish pixel-wise correspondences between support and query images. 
In addition, we design a Multi-Scale Feature Enhancement (MSFE) module that incorporates the Convolutional Block Attention Module (CBAM)~\cite{2018CBAM} and Atrous Spatial Pyramid Pooling (ASPP) module~\cite{chen2017rethinking} to integrate multi-scale features with prior mask information effectively.

The main contributions of this paper are summarized as follows:
\begin{itemize}
    \item[$\bullet$]
    We develop an end-to-end network for few-shot crack segmentation under low-light conditions, enabling direct extraction of crack regions from low-light images with limited training samples.

    \item[$\bullet$]
    We design a cross-similarity prior mask generation module to transform semantically rich high-level features into query prior masks by computing similarities between RGB and reflectance representations, thereby improving crack localization and structural characterization.

    \item[$\bullet$]
    We design an attention-based multi-scale feature enhancement module that dynamically fuses query features, support features and prior mask, suppressing background interference while preserving foreground details.

    \item[$\bullet$]
    We conduct experiments on multiple real-world and synthetic low-light crack datasets, demonstrating that the proposed method consistently outperforms state-of-the-art approaches.
\end{itemize}
\section{Related Work}
This section reviews recent advances in computer vision–based low-light crack segmentation and summarizes representative few-shot learning strategies for image segmentation.

\subsection{Low-light Crack Segmentation}

Crack segmentation is inherently a pixel-level binary classification problem. 
Image-based crack segmentation not only enables accurate characterization of crack morphology but also facilitates precise quantitative measurement of geometric attributes, such as length, width and area.

Similar to many visual recognition tasks, deep learning techniques have achieved remarkable progress in crack segmentation. 
However, most existing deep learning methods~\cite{8517148,9525150,10611660,liu2021crackformer,kang2022efficient} are developed under the assumption of high-quality input images and therefore exhibit significant performance degradation under adverse imaging conditions.
In real-world applications, image degradations such as low illumination are inevitable, robust segmentation under such conditions is imperative for applications like infrastructure inspection and autonomous driving.

To address these challenges, a variety of approaches have been proposed. 
Dai et al.~\cite{guo2023pavement} utilized an intermediate domain to adapt semantic models trained on daytime images to nocturnal settings, while Fan et al.~\cite{8569387} specifically targeted shadowed crack images by introducing a two-step shadow removal process. 
Other methods~\cite{10153729,8813888,Sun2019SeeCA} rely on additional image enhancement models to preprocess low-light images. 
More recent approaches~\cite{10611660,8803299,9697984} aim to bypass explicit enhancement by leveraging Retinex theory, adversarial learning and contrastive learning. 
These methods directly segment low-light crack images, 
thereby simplifying the overall pipeline and significantly reducing training complexity and computational cost.

\subsection{Few-Shot Segmentation}

Current few-shot semantic segmentation (FSS) methods can generally categorized into three classes: optimization-based, augmentation-based and metric learning approaches. Optimization-based methods~\cite{10670534,tian2019differentiablemetalearningmodelfewshot,boudiaf2021fewshotsegmentationmetalearninggood} employ gradient update strategies to mitigate data bias and enhance model generalization. 
Augmentation-based methods~\cite{lu2021simplerbetterfewshotsemantic,mehrotra2017generativeadversarialresidualpairwise,xian2019fvaegand2featuregeneratingframework} address data scarcity by synthesizing additional training samples.

Our work adopts a prototype-based metric learning paradigm, which derives representative class prototypes from support set features. 
These prototypes are subsequently compared with query image features using similarity metrics to predict pixel-wise segmentation masks for the query images.

The prototype-based paradigm is widely favored for its generalization ability, noise robustness, and computational efficiency, driving significant advances in recent research. Fan et al.~\cite{kim2024shotsegmentationrevealscompositional} introduced a self-supporting network to handle appearance variations, while Wang et al.'s PANet~\cite{fan2022selfsupportfewshotsemanticsegmentation} generates consistent class prototypes by leveraging both support and query features. 
Meanwhile, Liu et al.'s CRNet~\cite{wang2020panetfewshotimagesemantic} jointly predicts masks for support and query images through a cross-referencing mechanism that weights features for representation consistency. 
Furthermore,~\cite{liu2020crnetcrossreferencenetworksfewshot,yang2020prototypemixturemodelsfewshot,gairola2020simpropnetimprovedsimilaritypropagation} enhance cross-distribution adaptability by employing multiple prototypes to model local image features and associate distinct regions, thereby improving generalization performance.

\begin{figure*}[!t]
    \centering
    \includegraphics[width=\textwidth]{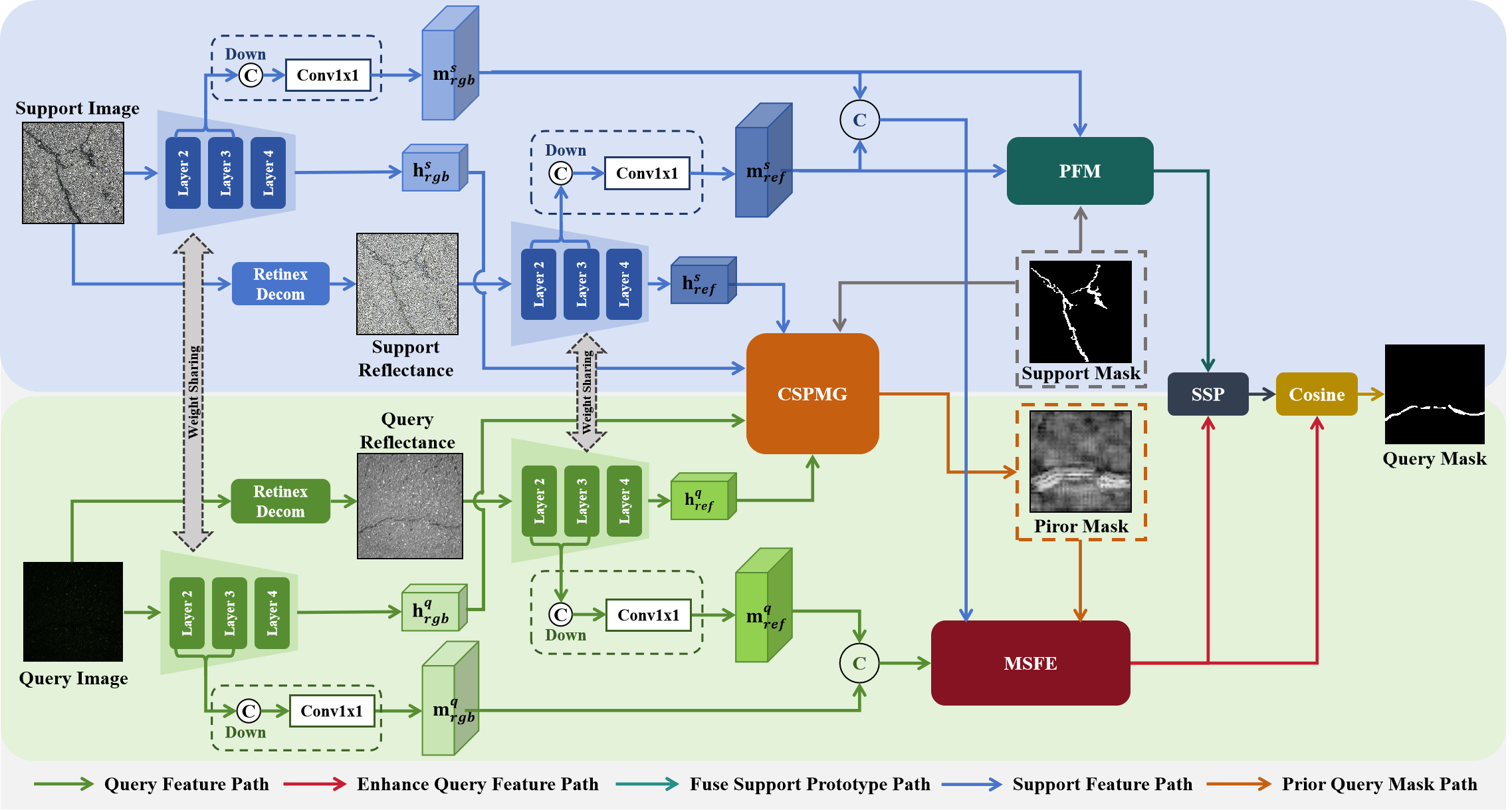}
    \caption{
        Architecture overview of the proposed network. The framework consists of five parts: 
        (a) dual feature extractors processing RGB images and corresponding reflectance component; 
        (b) a Cross-Similarity Prior Mask Generation (CSPMG) module hat produces query prior masks from high-level feature; 
        (c) a Multi-Scale Feature Enhancement (MSFE) module that refines query features through support-guided attention; 
        (d) a Prototype Fusion Module (PFM) that adaptively updates support prototype;
        (e) a Self-Support Prototype (SSP) module that further updates the support prototype.
    } \label{fig:image2}
\end{figure*}

\section{Methodology}
In this section, we detailed the architecture of our proposed few-shot crack segmentation network, which is designed to segment crack regions under low-light conditions using limited annotated training samples.


\subsection{Task Description}
Following the episodic meta-learning paradigm for few-shot semantic segmentation~\cite{Weakly-supervised}, our model is trained on a set of base classes $\mathcal{C}_{\text{train}}$ composed of well-illuminated crack images.
Each training episode consists of a support set $\mathcal{S} = \{(\mathbf{I}^{s}_i, \mathbf{M}^{s}_i)\}^{\text{K}}_{i=1}$ and a query set $\mathcal{Q} = \{(\mathbf{I}^{q}, \mathbf{M}^{q})\}$, where $\mathbf{I}^{*}$ denotes an RGB image and $\mathbf{M}^{*}$ represents its corresponding binary mask.
After training, the model is evaluated on a set of unseen novel classes $\mathcal{C}_{\text{test}}$, which contain only low-light crack images, satisfying $\mathcal{C}_{\text{train}} \cap \mathcal{C}_{\text{test}} = \varnothing$.

In the general N-way-K-shot segmentation setting, each episode involves N foreground classes, with K annotated support samples provided for each class.
In this work, we focus on the 1-way-K-shot scenario, which is particularly suited for crack segmentation, since cracks constitute a single foreground category against a complex background.
We evaluate our method under the 1-shot and 5-shot settings.
In the 1-shot case, the model predicts the query mask using one annotated support image, whereas in the 5-shot case, five support images are utilized.

\subsection{The Overall Architecture}
As illustrated in Fig.~\ref{fig:image2}, our segmentation pipeline processes a query image under 1-shot setting through the following stages:

\textbf{Retinex Decomposition and Feature Extraction}. 
Given a support–query image pair $\mathbf{I}^q$ and $\mathbf{I}^s$, we first decompose RGB images into their corresponding reflectance components $\mathbf{R}^s$ and $\mathbf{R}^q$ using a pretrained retinex decomposition network~\cite{Chen2018Retinex}. 
To extract illumination-aware and illumination-invariant representations in parallel, we employ two pairs of ResNet backbones pretrained on ImageNet-1K dataset~\cite{he2016deep} with shared weights, donated as $f_{rgb}(\cdot)$ and $f_{ref}(\cdot)$ for processing RGB images and reflectance components.
For the $i$-th convolutional block of the ResNet backbone, where $i\in\{1,2,\dots,5\}$, the extracted support features $\{\mathbf{F}_{rgb}^{s,i},\mathbf{F}_{ref}^{s,i}\}$ and query features $\{\mathbf{F}_{rgb}^{q,i},\mathbf{F}_{ref}^{q,i}\}$  are defined as:
\begin{equation}
\left\{
\begin{array}{lr}
     \mathbf{F}_{rgb}^{s,i},\mathbf{F}_{rgb}^{q,i} = f^i_{rgb}(\mathbf{I}^s,\mathbf{I}^q) &  \\
     \mathbf{F}_{ref}^{s,i},\mathbf{F}_{ref}^{q,i} = f^i_{ref}(\mathbf{R}^s,\mathbf{R}^q) & 
\end{array}
\right.
\end{equation}


High-level features tend to encode stronger semantic and category-specific information than mid-level features, which makes them more prone to overfitting seen categories and thus less transferable to unseen ones~\cite{PirorGuide}.
Accordingly, we organize the extracted features into mid-level features $\mathbf{m}^{\varepsilon}=\{\mathbf{m}^{\varepsilon}_{rgb},\mathbf{m}^{\varepsilon}_{ref}\}$ and high-level features $\mathbf{h}^{\varepsilon}=\{\mathbf{h}^{\varepsilon}_{rgb},\mathbf{h}^{\varepsilon}_{ref}\}$, where $\varepsilon\in \{s,q\}$ respectively denotes support and query sets.

Specifically, the mid-level features $\mathbf{m}^{\varepsilon}\in \mathbb{R}^{C_m\times H_m \times W_m}$ are obtained by fusing the outputs of the 2-th and 3-th blocks through a downsampling module $D(\cdot)$:
\begin{equation}
    \begin{aligned}
    \mathbf{m}^\varepsilon & = \{\mathbf{m}^\varepsilon_{rgb},\mathbf{m}^\varepsilon_{ref}\} \\
                & = \{D(\mathbf{F}_{rgb}^{\varepsilon,2}, \mathbf{F}_{rgb}^{\varepsilon,3}), D(\mathbf{F}_{ref}^{\varepsilon,2}, \mathbf{F}_{ref}^{\varepsilon,3})\}
    \end{aligned}
\end{equation}
 
Here, $D(\cdot)$ performs channel-wise concatenation followed by a $1\times1$ convolution for dimensionality alignment.

In contract, the high-level features $\mathbf{h}^{\varepsilon} \in \mathbb{R}^{C_h\times H_h \times W_h}$ are directly extracted from the 5-th block:
\begin{equation}
    \mathbf{h}^\varepsilon = \{\mathbf{h}^\varepsilon_{rgb},\mathbf{h}^\varepsilon_{ref}\} = \{\mathbf{F}_{rgb}^{\varepsilon,5}, \mathbf{F}_{ref}^{\varepsilon,5}\}
\end{equation}

\textbf{Support Prototype Vectors Fusion}. 
The mid-level support features $\{\mathbf{m}^s_{rgb},\mathbf{m}^s_{ref}\}$ are aggregated into the support prototype vectors $\{\bm{P}_s,\bm{P}_{sr}\}$ using Masked Average Pooling (MAP)~\cite{SSP} guided by the support mask $\mathbf{M}^s$.
To effectively integrate complementary from RGB and reflectance components, we adopt the Prototype Fusion Module (PFM) proposed in CrackNex~\cite{CrackNex}, which produces a unified support prototype vector $\bm{P}'_s \in \mathbb{R}^{1\times 1\times C_m}$:
\begin{equation}
    \bm{P}'_s = \text{PFM}(\mathbf{m}^s_{rgb}, \mathbf{m}^s_{ref}, \mathbf{M}^s)
\end{equation}

\textbf{Prior Query Mask Generation}. 
To provide explicit spatial guidance for query segmentation, we introduce the CSPMG module to cross-image similarities between the support and query high-level features and produces a prior mask $\mathbf{M}^{q}_{prior}$, which encodes coarse crack localization cues for the query image. The detailed design of the CSPMG module is described in Section~\ref{Section:CSPMG}.
\begin{equation}
    \mathbf{M}^{q}_{prior} = \text{CSPMG}(\mathbf{h}^s, \mathbf{h}^q, \mathbf{M}^s)
\end{equation}

\textbf{Query Feature Enhancement}.
The MSFE module further refines the query representation by jointly leveraging the mid-level support features and the generated prior mask.
Specifically, the MSFE module integrates these complementary cues to produce enhanced query features
$\mathbf{m}_{fuse}^q \in \mathbb{R}^{C_m \times H_m \times W_m}$. 
The architectural details of the MSFE module are described in Section~\ref{Section:MSFE}.
\begin{equation}
    \mathbf{m}^q_{fuse} = \text{MSFE}(\mathbf{m}^s_c, \mathbf{m}^q_c, \mathbf{M}^s, \mathbf{M}^{q}_{\text{prior}})
\end{equation}
where $\mathbf{m}^s_c$ and $\mathbf{m}^q_c$ respectively denote the channel-wise concatenations of $\{\mathbf{m}^s_{rgb},\mathbf{m}^s_{ref}\}$ and $\{\mathbf{m}^q_{rgb},\mathbf{m}^q_{ref}\}$.

\textbf{Prototype Metric and Mask Prediction}. 
A metric-based matching operation is conducted between the enhanced query features $\mathbf{m}_{fuse}^q$ and the fused support prototype vector $\bm{P}_s'$.

Updated support prototype $\bm{P}_s'$ is then fed into
the Self-Support Prototype (SSP) module with enhanced query features $\mathbf{m}^q_{fuse}$. 
The output is augmented prototype $\bm{P}_s''\in \mathbb{R}^{1\times1\times C}$.
The resulting cosine similarity map is then transformed into the final segmentation prediction $\mathbf{\hat{y}} \in \mathbb{R}^{1 \times H \times W}$ via a softmax activation function:
\begin{equation}
    \mathbf{\hat{y}} = \text{Softmax}(\text{cosine}(\bm{P}_s'', \mathbf{m}^q_{fuse}))
\end{equation}


\subsection{Cross-Similarity Prior Mask Generation Module}
\label{Section:CSPMG}
In a high-dimensional feature space, each pixel is represented as a feature vector extracted by the feature extraction backbone network. 
These vectors encode rich local semantic information and their distances between each other reflect both feature similarity and pixel-level semantic correspondence across images.

\begin{figure*}[!t]
    \centering
    \includegraphics[width=1\textwidth]{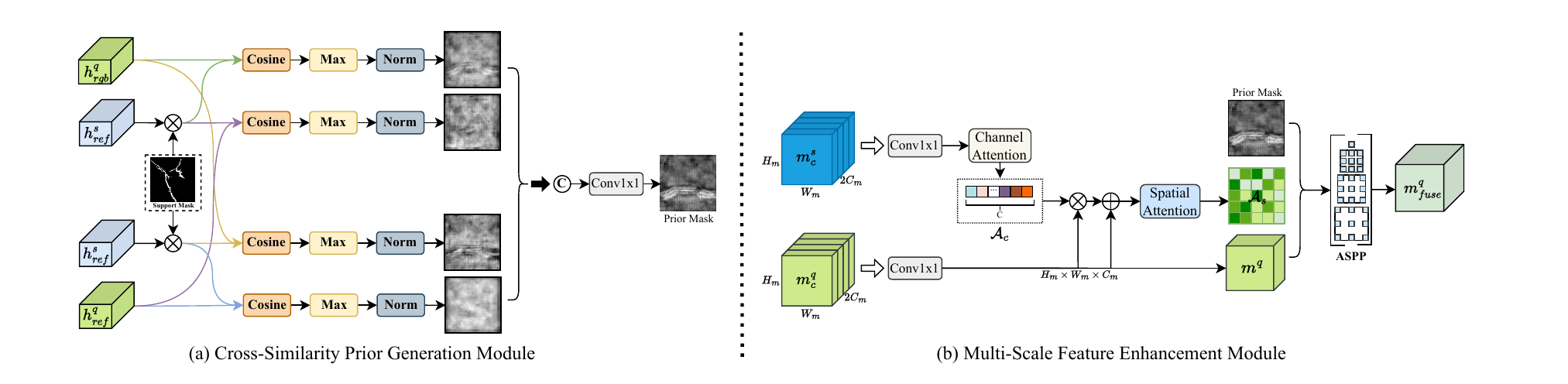}
    \caption{
        (a) The architecture of the Cross-Similarity Prior Mask Generation (CSPMG) module;
        (b) The architecture of the Multi-Scale Feature Enhancement (MSFE) module.
    } \label{fig:image3}
\end{figure*}

To explicitly model the correspondence between crack regions in the support and query images, we propose a Cross-Similarity Prior Mask Generation (CSPMG) module and the architecture of this module is shown in Fig.~\ref{fig:image3} (a).
This module computes dense pixel-wise cosine similarities between support and query features, producing a query prior mask that captures coarse structural and spatial information of crack regions.
The generated prior mask subsequently serves as an explicit guidance signal for query feature enhancement.

The CSPMG module takes as input the high-level features of the support and query images, respectively denoted as $\mathbf{h}^s = \{\mathbf{h}^s_{rgb},\mathbf{h}^s_{ref}\}$ and $\mathbf{h}^q = \{\mathbf{h}^q_{rgb},\mathbf{h}^q_{ref}\}$. 
To suppress background regions that are irrelevant to cracks, the high-level support features $\mathbf{h}^s$ are masked using the binary support mask $\mathbf{M}^s \in \{0,1\}^{H \times W}$:
\begin{equation}
\mathbf{h}^{s}_m = \mathbf{h}^s \odot \delta(\mathbf{M}^s)
\end{equation}
where $\odot$ denotes the Hadamard product and $\delta(\cdot):\mathbb{R}^{H\times W}\rightarrow \mathbb{R}^{H_h\times W_h}$ represents bilinear interpolation for spatial alignment.

For pixel-wise similarity computation, the feature maps are vectorized along the channel dimension. 
The cosine similarity $d_{ij}$ between the query feature vector $\boldsymbol{e}_i^q \in \mathbf{h}^q$ and the masked support feature vector $\boldsymbol{e}_j^s \in \mathbf{h}^{s}_m$ is defined as:
\begin{equation}
d_{ij} = \text{cosine}(\boldsymbol{e}_i^q, \boldsymbol{e}_j^s) = \frac{(\boldsymbol{e}_i^q)^\top \boldsymbol{e}_j^s}{\Vert \boldsymbol{e}_i^q \Vert \Vert \boldsymbol{e}_j^s \Vert}
\end{equation}

Here, $i,j\in \{1,2,\dots,H_hW_h\}$ index spatial locations in the query and support feature maps.

Exploiting the intra-class clustering property of crack features in the embedding space, we compute the average similarity of each query pixel with respect to all crack pixels in the support image:
\begin{equation}
v_i = \frac{1}{N}\sum_{j=1}^{N} \cos(\boldsymbol{e}_i^q, \boldsymbol{e}_j^s)
\end{equation}

This process yields a similarity vector $\boldsymbol{v} = \{v_1, v_2,\dots, v_{N}\}$ for the query image, where $N = H_h \times W_h$ denotes the size of feature maps.
Reshaping $\boldsymbol{v}\in \mathbb{R}^{1\times H_hW_h}$ into a similarity matrix $\mathbf{V} \in \mathbb{R}^{H_h \times W_h}$, we then normalize it to the range between 0 and 1:
\begin{equation}
\mathbf{M} = \frac{\mathbf{V} - \min(\mathbf{V})}{\max(\mathbf{V}) - \min(\mathbf{V}) + \mu}
\end{equation}
where $\mu=1\times10^{-6}$ is introduced to ensure numerical stability.

Since the support and query feature representations comprise both RGB and reflectance components, we further compute four prior masks $\mathbf{M}_{prior}=\{\mathbf{M}_{\alpha,\beta}\}$, where $\alpha,\beta\in\{rgb,ref\}$.

These modality-specific prior masks are then fused using a $1\times1$ convolution followed by a channel-wise softmax operation:
\begin{equation}\label{12}
\mathbf{M}_{prior}^q = \text{Softmax}(\text{Conv}(\text{Cat}(\mathbf{M}_{prior})))
\end{equation}

Here, $\text{Cat}(\cdot)$ indicates channel concatenation and $\mathbf{M}_{\text{prior}}^q \in \mathbb{R}^{H_h\times W_h}$.

\begin{figure}[ht]
    \centering
    \includegraphics[width=0.45\textwidth]{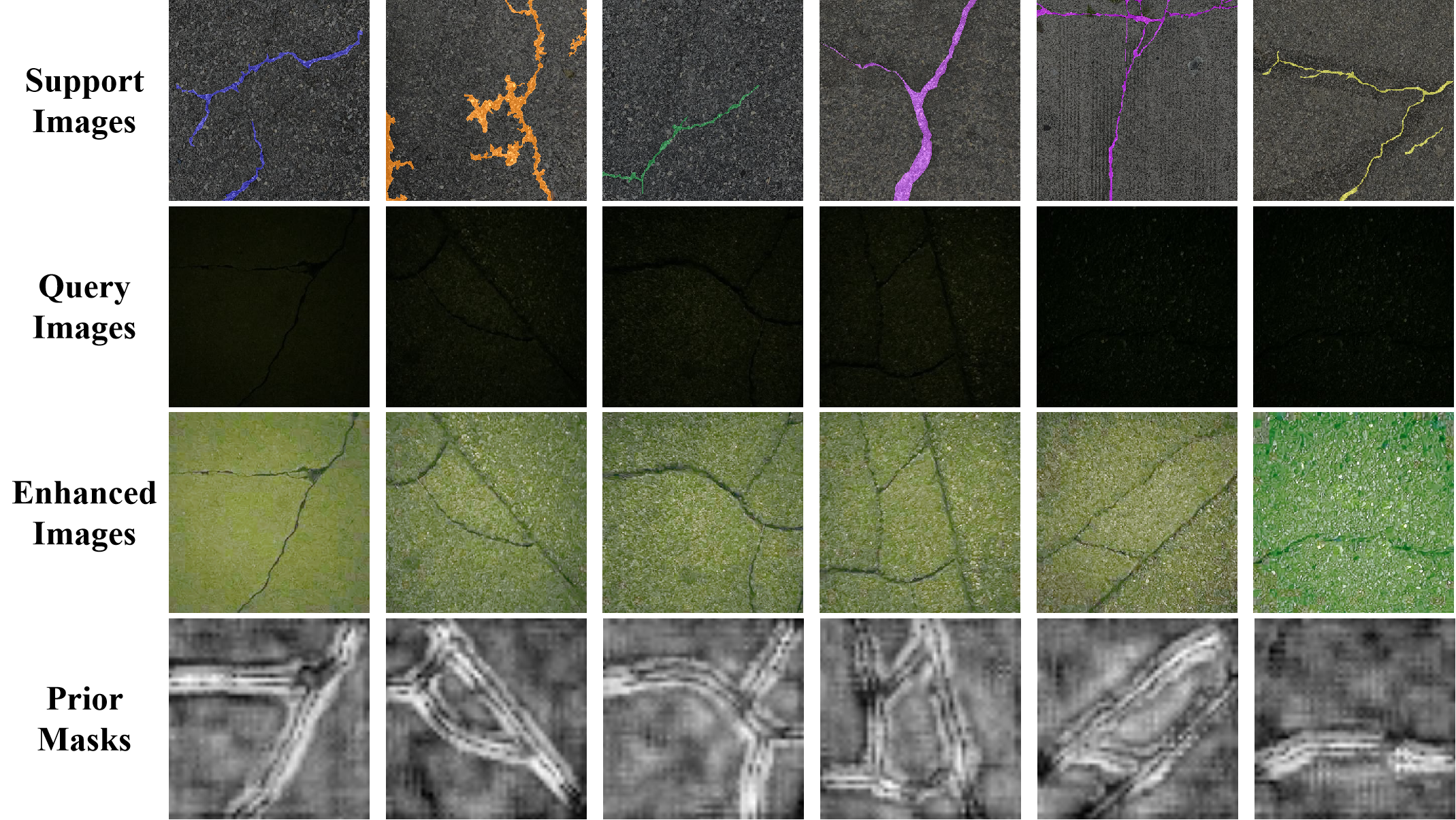}
    \caption{
        Prior mask generation and qualitative results on the LCSD dataset. 
        \textbf{Top}: Support images with crack regions masked.
        \textbf{Middle}: Query images and corresponding enhanced visualizations.
        \textbf{Bottom}: Generated prior masks highlighting regions of interest in the query images.
    } \label{fig:image4}
\end{figure}

The prior mask $\mathbf{M}_{prior}^q$ is upsampled to the original query image size for visualization. 
As shown in Fig.~\ref{fig:image4}, we display these results on six real-world low-light crack images from the LCSD dataset~\cite{CrackNex}, where the prior masks effectively highlight the probable crack regions.

\subsection{Multi-Scale Feature Enhancement Module}
\label{Section:MSFE}
In this section, we present the Multi-Scale Feature Enhancement (MSFE) module, which is designed to effectively integrate guidance from the support features and the generated prior mask into the query branch. 
As illustrated in Fig.~\ref{fig:image3} (b), the MSFE module employs both channel-wise and spatial attention mechanisms to adaptively refine query feature representations, thereby enhancing the model’s focus on regions relevant to target cracks and improving segmentation performance.

Firstly, to fuse multi-modal information and reduce feature dimensionality, we concatenate the mid-level RGB and reflectance features along the channel dimension and apply a $1 \times 1$ convolution for dimensionality:
\begin{equation}
    \mathbf{m}^\varepsilon_{c} = \text{Conv} \left( \text{Cat}(\mathbf{m}^\varepsilon_{rgb}, \mathbf{m}^\varepsilon_{ref}) \right)
\end{equation}
where $\mathbf{m}^\varepsilon_{c}\in \mathbb{R}^{C_m\times H_m \times W_m}$ donates the fused mid-level features, and $\varepsilon\in\{s,q\}$ indicates the support or query set.

To emphasize channels that are discriminative for crack-related patterns, we apply a channel attention mechanism to the fused support features. 
Specifically, Global Average Pooling (GAP) and Global Max Pooling (GMP)~\cite{2018CBAM} are used to capture global contextual information, followed by shared multi-layer perceptrons (MLPs) and a sigmoid activation function to generate channel-wise attention weights:
\begin{equation}
    \mathbf{A}_c = \sigma \left( \text{MLP}(\text{GAP}(\mathbf{m}^s_{c})) + \text{MLP}(\text{GMP}(\mathbf{m}^s_{c})) \right)
\end{equation}
where $\sigma(\cdot)$ denotes the sigmoid function and the channel attention weight $\mathbf{A}_c\in \mathbb{R}^{1\times 1\times C_m}$ is then used to adaptively recalibrate the query features:
\begin{equation}
    \mathbf{\widetilde{m}}^q_c = (1 + \mathbf{A}_c) \odot \mathbf{m}^q_c
\end{equation}



Next, a spatial attention mechanism is introduced to further refine the spatial distribution of the channel-enhanced query features. 
We aggregate channel-enhanced features $\mathbf{\widetilde{m}}^q$ using both GAP and GMP, concatenate the resulting spatial descriptors, and apply a $3 \times 3$ convolution followed by a sigmoid activation to obtain a spatial attention map $\mathbf{A}_s$:
\begin{equation}
    \mathbf{A}_s = \sigma \left( \text{Conv} \left( [\text{GAP}(\mathbf{\widetilde{m}}^q_c), \text{GMP}(\mathbf{\widetilde{m}}^q_c)] \right) \right)
\end{equation}
where $\mathbf{A}_s \in \mathbb{R}^{1 \times H_m \times W_m}$ highlights spatial locations that are likely to correspond to crack regions, complementing the channel-wise modulation.

Finally, we concatenate the prior mask $\mathbf{M}^q_{fuse}$, the spatial attention weight $\mathbf{A}_s$ and the channel-enhanced query features $\mathbf{\widetilde{m}}^q$, and feed them into an Atrous Spatial Pyramid Pooling (ASPP) module~\cite{chen2017rethinking}:
\begin{equation}
    \mathbf{m}^q = \text{ASPP}(\text{Cat}(\mathbf{\widetilde{m}}^q_c,\mathbf{M}_{prior}^q,\mathbf{A}_s))
\end{equation}

The ASPP module captures multi-scale contextual information, enabling robust adaptation to cracks with varying sizes and shapes.
This enhancement strategy jointly exploits global context, local structural cues, and explicit localization guidance from the prior mask, leading to more accurate segmentation.




\subsection{Loss Function}
We trained our model in a supervised manner using a combination of segmentation and auxiliary losses.
For the final query segmentation prediction, we adopt the binary cross-entropy (BCE) loss:
\begin{equation}
    \mathcal{L}_{seg} = \text{BCE}(\mathbf{\hat{y}},\mathbf{M}^q)
\end{equation}
where $\mathbf{\hat{y}}$ denotes the predicted segmentation mask and $\mathbf{M}_q \in \{0,1\}^{H\times W}$ is the ground truth label of the query image.
$\mathcal{L}_{seg}$ enforces consistency between the predicted mask and the ground-truth labels.

To explicitly supervise the quality of the generated prior mask, we further introduce a prior consistency loss:
\begin{equation}   
    \mathcal{L}_{prior} = \text{BCE}(\mathbf{M}^q_{prior},\mathbf{M}^q)
\end{equation}
which encourages the prior mask to provide accurate localization cues for crack regions in the query image.

Moreover, to facilitate stable optimization of the SSP module, we follow the self-support loss proposed in~\cite{SSP}, which enforces consistency between the learned support prototypes and the corresponding support features from both RGB and reflectance branches:
\begin{equation}
\begin{aligned}
\mathcal{L}_{ssp} = \text{BCE}(\text{cosine}(\bm{P}_s^{'},\mathbf{h}_{rgb}^{s}),\mathbf{M}^s) 
                    \\+ \text{BCE}(\text{cosine}(\bm{P}_s^{'},\mathbf{h}_{ref}^{s}),\mathbf{M}^s)
\end{aligned}
\end{equation}

The overall training objective is defined as a weighted sum of the above loss terms:
\begin{equation}
    \mathcal{L} = \lambda_1 \mathcal{L}_{seg} + \lambda_2 \mathcal{L}_{prior} + \lambda_3 \mathcal{L}_{ssp}
\end{equation}
where the weighting coefficients $\lambda_1$, $\lambda_2$ and $\lambda_3$ are respectively set to 0.1, 0.5 and 0.6.

\section{Experiments}
\subsection{Datasets}
To comprehensively evaluate the effectiveness of the proposed model under low-light conditions, we conduct experiments on both real-world and synthetic low-light crack segmentation datasets.

1) \textbf{LCSD}: 
To evaluate our model under real-world low-light conditions, we adopt the LCSD dataset introduced by Yao et al.~\cite{CrackNex}.
LCSD is collected on the Lehigh University campus and consists of 102 well-illuminated crack images used for training and 41 low-light crack images reserved for testing.
All images are captured using a first-generation iPad Pro and resized to $400\times 400$ pixels for computational efficiency.
Each image is annotated at the pixel level, yielding binary ground-truth masks for crack 

2) \textbf{llCrackSeg9k}:  
CrackSeg9k~\cite{CrackSeg9k} is a widely used benchmark dataset for crack segmentation.
Following prior works~\cite{CrackNex}, we select 9000 images as the training set and an additional 1500 images as the test set.
Since the original test images are captured under normal lighting conditions, we employ Restormer pretrained on the LDIS dataset~\cite{Restormer} to transform them into synthetic low-light images, thereby constructing the llCrackSeg9k benchmark for evaluating low-light crack segmentation performance.

\subsection{Implementation Details}
For the backbone networks, we adopt ResNet-50 and ResNet-101~\cite{he2016deep} pretrained on ImageNet-1K dataset~\cite{chen2017rethinking}.

The entire framework is optimized using the AdamX optimizer~\cite{adamx} with an initial learning rate of $1\times 10^{-3}$, which is decayed by a factor of 10 every 2,000 iterations. 
The model is trained for 6000 iterations with a batch size of 4 on a single NVIDIA RTX 3090 GPU with 24 GB memory.

During training, both images and their corresponding masks are augmented using random horizontal flipping, while evaluation is conducted on the original images without augmentation.

For a fair comparison, we further evaluate several state-of-the-art (SOTA) models~\cite{MLC,SSP,CrackNex} on the LCSD and llCrackSeg9k datasets under the same experimental protocol.

\subsection{Quantitative Results}
We adopt mean Intersection-over-Union (mIoU) as the evaluation metric and report results under both 1-shot and 5-shot settings.
Quantitative comparisons are conducted on both the llCrackSeg9k and LCSD benchmarks.
As summarized in Table~\ref{tab:table1}, our method consistently achieves superior performance among several SOTA approaches, including MLC~\cite{MLC}, SSP~\cite{SSP}, and CrackNex~\cite{CrackNex}, using their default configurations.

On the llCrackSeg9k dataset, our approach attains mIoU scores of 66.20 and 69.93 under the 1-shot and 5-shot settings, respectively, with the ResNet-50 backbone, and further improves to 68.20 and 71.40 with ResNet-101. These results demonstrate clear advantages over existing methods.

We further evaluate the proposed model on the LCSD dataset, which contains real-world low-light images. 
As shown in Table ~\ref{tab:table1}, our method achieves mIoU scores of 66.86 (1-shot) and 68.04 (5-shot) with ResNet-50, and 70.37 (1-shot) and 69.27 (5-shot) with ResNet-101. 
The consistent performance gains indicate strong robustness and generalization capability in practical low-light scenarios.

\begin{table}[ht]
    \caption{
        Baseline comparisons on the llCrackSeg9k and LCSD datasets in terms of mIoU$\uparrow$
    }
    \vspace{0.2cm}
    \normalsize
    \center
    \resizebox{0.5\textwidth}{!}{
        \begin{tabular}{c|c|ll|ll}
            \hline
                                    &                             & \multicolumn{2}{c|}{\textbf{llCrackSeg9k}}                                                                         & \multicolumn{2}{c}{\textbf{LCSD}}                                                                               \\ \cline{3-6} 
            \multirow{-2}{*}{models} & \multirow{-2}{*}{Backbone}  & \multicolumn{1}{c|}{1-shot}                                   & \multicolumn{1}{c|}{5-shot}                     & \multicolumn{1}{c|}{1-shot}                                   & \multicolumn{1}{c}{5-shot}                      \\ \hline
            MLC~\cite{MLC}                       &                             & 56.54                                                         & 58.72                                           & 55.48                                                         & 57.41                                           \\
            SSP~\cite{SSP}                       &                             & 60.42                                                         & 64.25                                           & 56.41                                                         & 63.30                                           \\
            CrackNex~\cite{CrackNex}                  &                             & 63.00                                                         & 69.66                                           & 63.68                                                         & 65.17                                           \\
            \textbf{Ours}             & \multirow{-4}{*}{ResNet-50}  & \cellcolor[HTML]{C0C0C0}\textbf{66.20} $\uparrow$             & \cellcolor[HTML]{C0C0C0}\textbf{69.93} $\uparrow$                        & \cellcolor[HTML]{C0C0C0}\textbf{66.86} $\uparrow$             & \cellcolor[HTML]{C0C0C0}\textbf{68.04} $\uparrow$  \\ \hline
            MLC~\cite{MLC}                       &                             & 56.73                                                         & 62.99                                           & 57.18                                                         & 58.11                                           \\
            SSP~\cite{SSP}                       &                             & 56.45                                                         & 65.29                                           & 56.61                                                         & 63.16                                           \\
            CrackNex~\cite{CrackNex}                  &                             & 65.90                                                         & 70.59                                           & 66.10                                                         & 68.82                                           \\
            \textbf{Ours}             & \multirow{-4}{*}{ResNet-101} & \cellcolor[HTML]{C0C0C0}\textbf{68.20} $\uparrow$             & \cellcolor[HTML]{C0C0C0}\textbf{71.40} $\uparrow$ & \cellcolor[HTML]{C0C0C0}\textbf{70.37} $\uparrow$             & \cellcolor[HTML]{C0C0C0}\textbf{69.27} $\uparrow$ \\ \hline
        \end{tabular}
    }
    \label{tab:table1}
    \vspace{-0.2cm}
\end{table}

\subsection{Qualitative Results}
\begin{figure}[ht]
    \centering
    \includegraphics[width=0.5\textwidth]{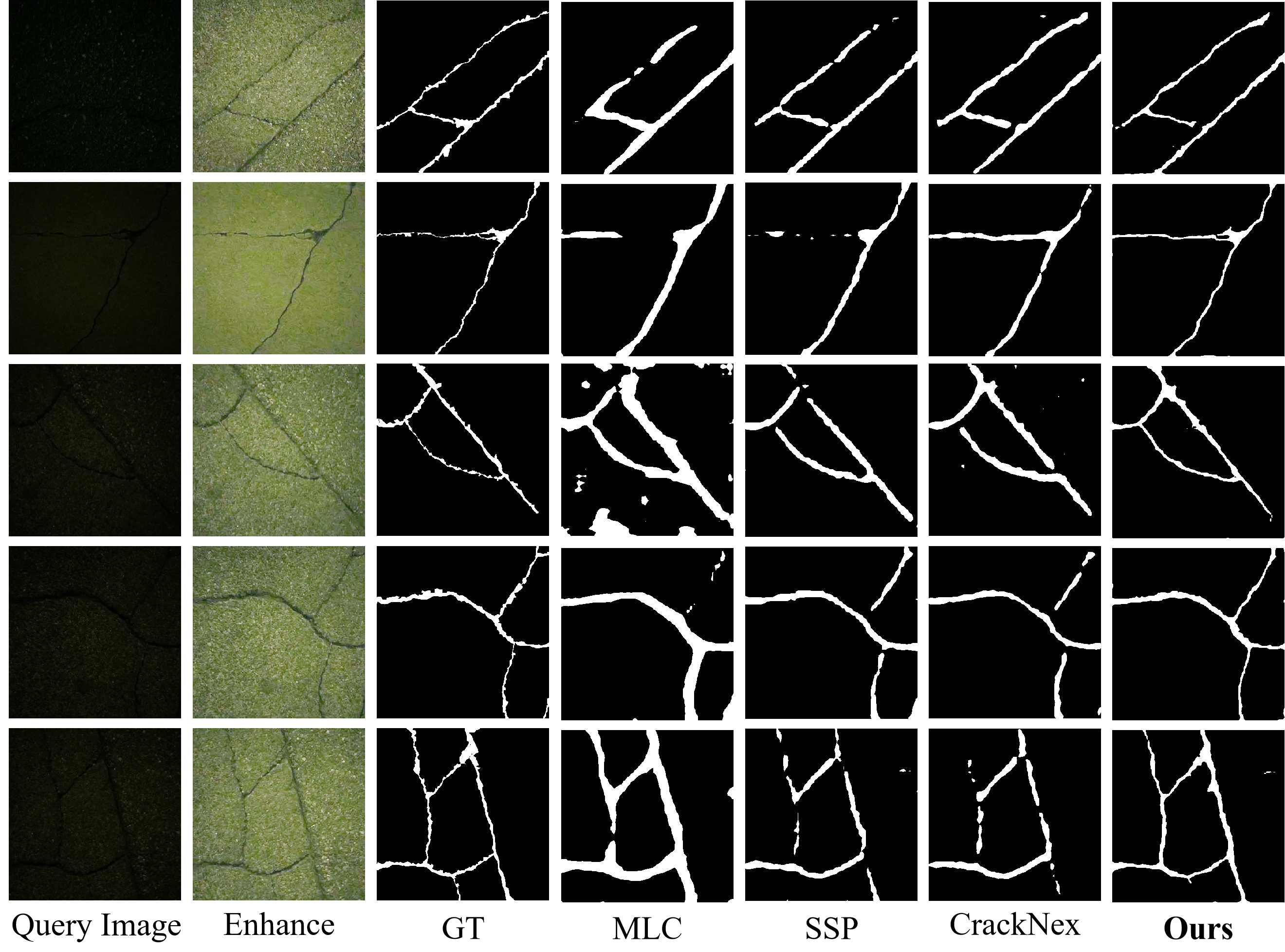}
    \caption{
            Qualitative comparison of segmentation results on the LCSD dataset under the 1-shot setting. From left to right: (a) query image (contrast-enhanced for visualization), (b) ground truth, (c) MLC~\cite{MLC}, (d) SSP~\cite{SSP}, (e) CrackNex~\cite{CrackNex}, and (f) our model. 
            }\label{fig:image5}
\end{figure}

To provide qualitative insights into segmentation performance, we visualize representative results on the LCSD dataset under the 1-shot setting, as shown in Fig.~\ref{fig:image5}. 

With three SOTA approaches, namely SSP~\cite{SSP}, MLC~\cite{MLC}, and the recently proposed CrackNex~\cite{CrackNex}, all evaluated under their default configurations.

As shown in Fig.~\ref{fig:image5}, our method produces segmentation masks with clearer boundaries and improved spatial continuity. 
In challenging low-light regions with severe noise or low contrast, our method more effectively suppresses false positives while preserving fine-grained crack structures.

\subsection{Ablation Study}
We further conduct an ablation study to evaluate the effectiveness of each component in our proposed model on the LCSD dataset using the ResNet-101 backbone.
Specifically, we evaluate four components: 
(i) baseline architecture, 
(ii) utilizing CSPMG module, 
(iii) designing MSFE module and 
(iv) utilizing SSP module.

\begin{table}[ht]
    \caption{
        Ablation study of adding different components on LCSD dataset
    }
    \vspace{0.2cm}
    \label{tab:table2}
    \resizebox{1\linewidth}{!}{
        \begin{tabular}{cccc|cc}
        \hline
        \multirow{2}{*}{CSPMG Module} & \multirow{2}{*}{MSFE Module} & \multirow{2}{*}{PEM Module} & \multirow{2}{*}{SSP Module} & \multicolumn{2}{c}{mIOU} \\
                                    &                             &                             &                             & 1-shot      & 5-shot     \\ \hline
                                    &                             &                             &                             & 68.05       & 68.71      \\
        \ding{52}                   &                             &                             &                             & 68.20       & 68.14      \\
        \ding{52}                   & \ding{52}                   &                             &                             & 68.53       & 69.94      \\
        \ding{52}                   & \ding{52}                   & \ding{52}                   &                             & 68.57       & 67.14      \\
        \ding{52}                   & \ding{52}                   & \ding{52}                   & \ding{52}                   & 70.37       & 69.27      \\ \hline
        \end{tabular}
    }
    \vspace{-0.2cm}
\end{table}

\begin{figure}[ht]
    \centering
    \includegraphics[width=0.5\textwidth]{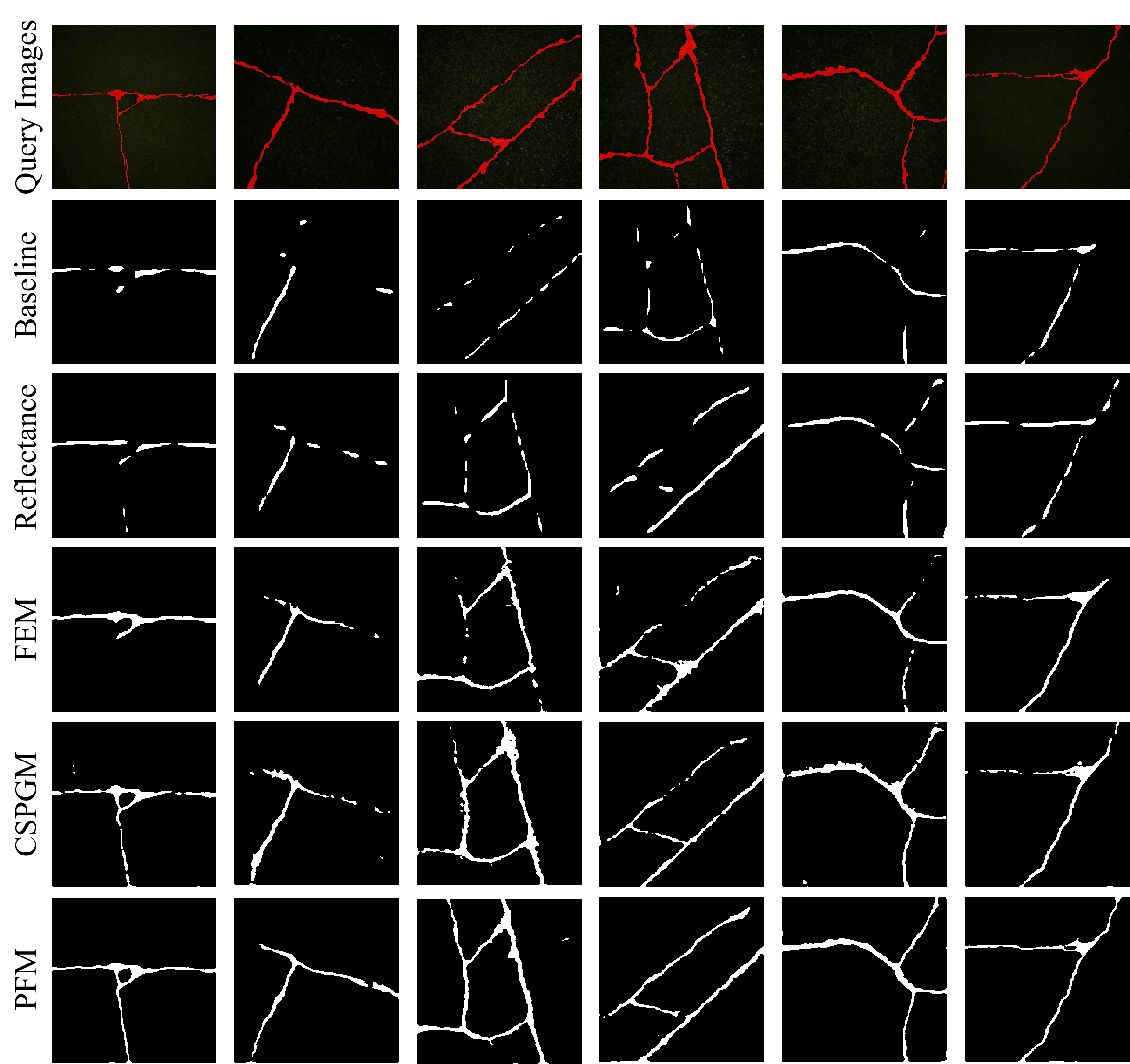}
    \caption{
            Visual results of the 1-shot ablation study with ResNet-101 backbone, from top to bottom, obtained by incrementally adding the corresponding modules to the baseline model. 
            }
    \label{fig:image6}
\end{figure}

The results in the Table~\ref{tab:table2} show that each individual component contributes to performance improvements, while the full model achieves the best results under both 1-shot and 5-shot settings. 
This confirms that the proposed modules are complementary and jointly enhance segmentation performance. 
Additional qualitative ablation results are visualized in Fig.~\ref{fig:image6}, further illustrating the progressive refinement achieved by integrating each module.


\subsection{Conclusion}
In this paper, we present a few-shot crack segmentation framework designed for low-light scenarios, addressing the challenges of limited annotated data and degraded visual features.
Formulated under a 1-way-$K$-shot setting, the proposed dual-branch prototypical metric network incorporates reflectance information derived from Retinex-based decomposition to learn illumination-invariant representations, thereby improving robustness under severe lighting variations.

To further enhance segmentation performance, we introduce a cross-similarity prior mask generation module and a multi-scale query feature enhancement module, which jointly provide spatial priors and strengthen query feature representations.
Extensive experiments on both synthetic and real-world low-light crack datasets demonstrate that our method consistently outperforms existing SOTA few-shot segmentation approaches, validating its effectiveness and generalization capability in challenging low-light environments.
\section*{Acknowledgment}


{\small
\bibliographystyle{reference/ieee_fullname}
\bibliography{reference/references}
}

\end{document}